\documentclass[lettersize,journal]{IEEEtran}
\usepackage{amsmath,amsfonts}
\usepackage{algorithmic}
\usepackage{algorithm}
\usepackage{array}
\usepackage[caption=false,font=normalsize,labelfont=sf,textfont=sf]{subfig}
\usepackage{textcomp}
\usepackage{stfloats}
\usepackage{url}
\usepackage{verbatim}
\usepackage{graphicx}
\usepackage{cite}
\usepackage{xcolor}
\usepackage{multirow}
\usepackage{booktabs}
\usepackage{color}
\usepackage{hyperref}

\begin{document}

\title{MinkUNeXt: Point Cloud-based Large-scale Place Recognition using 3D Sparse Convolutions}

\author{J.J. Cabrera$^{1}$, A. Santo$^{1,2}$, A. Gil$^{1}$, C. Viegas$^{3}$ and L. Payá$^{1}$ %
\thanks{$^{1}$Institute for Engineering Research (I3E) Email: \{juan.cabreram, a.santo, arturo.gil, lpaya\}@umh.es}%
\thanks{$^{2}$Valencian Graduate School and Research Network for Artificial Intelligence.}%
\thanks{$^{3}$Univ of Coimbra, ADAI, Department of Mechanical Engineering, Rua Luís Reis Santos, Pólo II, 3030-788 Coimbra, Portugal. Email: carlos.viegas@uc.pt }%
\thanks{979-8-3503-0704-7/23/\$31.00 \copyright 2023 European Union}
}



\maketitle

\begin{abstract}
This paper presents MinkUNeXt, an effective and efficient architecture for place-recognition from point clouds entirely based on the new 3D MinkNeXt Block, a residual block composed of 3D sparse convolutions that follows the philosophy established by recent Transformers but purely using simple 3D convolutions. Feature extraction is performed at different scales by a U-Net encoder-decoder network and the feature aggregation of those features into a single descriptor is carried out by a Generalized Mean Pooling (GeM). The proposed architecture demonstrates that it is possible to surpass the current state-of-the-art by only relying on conventional 3D sparse convolutions without making use of more complex and sophisticated proposals such as Transformers, Attention-Layers or Deformable Convolutions. A thorough assessment of the proposal has been carried out using the Oxford RobotCar and the In-house datasets. As a result, MinkUNeXt proves to outperform other methods in the state-of-the-art.
\end{abstract}


\section{Introduction}
\IEEEPARstart{I}{n} many applications, mobile robots must perform autonomous navigation in a specific environment. As it moves, the robot should be able to recognize or identify different areas of the environment. This action is equivalent to finding a correspondence between its current sensor observations and a part of the stored database. This ability is commonly denoted as place recognition. In order to speed this process, frequently, authors have concentrated on describing some parts of the environment by means of an invariant descriptor. In this way, the robot should be able to recognize a part of the environment by finding the descriptor in the database that most ressembles the descriptor associated to its current observations. The concept of place recognition is of uttermost importance in tasks such as localization, mapping and navigation.

\begin{figure}[t]
\centering
\includegraphics[width=0.5\textwidth]{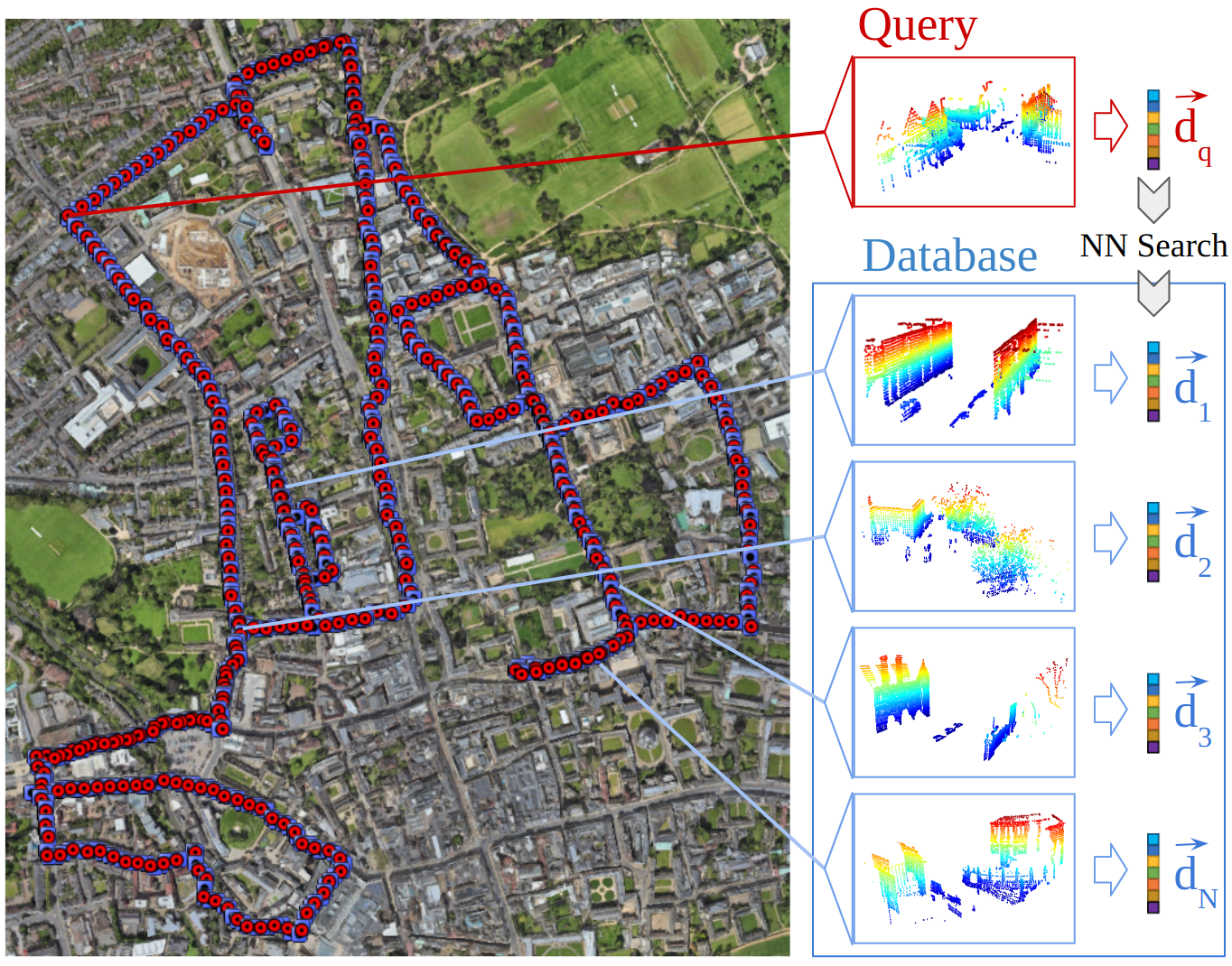}
\caption{Point cloud-based place recognition. Each query point cloud (red) is embedded into a global descriptor which is compared with the descriptors from the database point clouds (blue) by means of a Nearest Neighbour Search.}
\label{fig_place_recognition}
\end{figure}

\vspace{0.5cm}
Place recognition and robot localization are two closely related concepts. Place recognition centers on the description of the current robot observations in a way that allows the robot to identify different locations in the map. Thus, place recognition focuses on the extraction and codification of relevant features found in the robot query observation in such a way that they can be compared to previously stored data (Fig. \ref{fig_place_recognition}). Similarly, robot localization refers to the act of estimating the position and orientation of the robot within a known map. In this way, given a map of the robot, conformed by a series of submaps, a common process to carry out the global localization of the robot could consist of two phases \cite{yin2018locnet}: a) rapidly finding a submap within the global database using the feature descriptors (place recognition) and b) performing a fine estimation of the position and orientation of the robot in that submap (robot localization). A similar technique is proposed in \cite{kim2018scancontext}, where the descriptor is computed from LiDAR measurements. Next, a handcrafted descriptor is employed to rapidly retrieve some areas of interest in the map. The final localization step, based on the ICP (Iterative Closest Point) algorithm is able to compute the position and orientation within the submap.  

\vspace{0.5cm}
To date, place recognition has been performed with different types of sensors: visual cameras \cite{leyva2019placeSiamese}, laser \cite{himstedt2014large}, LiDAR \cite{komorowski2022improving} and Radar \cite{komorowski2021radar} using different types of techniques. For example, place recognition has been extensively solved by means of techniques based on the Bag of Words algorithm using images \cite{Tang2019bow, arshad2023bow}.

\vspace{0.5cm}
During the last few years LiDAR sensors have lowered in price and weight, while increasing in resolution. Therefore, LiDAR sensors permit obtaining a large number of precise measurements from the environment that define its shape and structure. Being a self-illuminated sensor, it is insensitive to changes in natural light, thus directly applicable to outdoor applications. In consequence, several new potential applications of LiDAR sensors in the area of mobile robotics have emerged and it is therefore necessary to focus on methods that achieve a robust description of the scene. In the literature, so far, we can find: a) Classical techniques based on a handcrafted description of LiDAR data to generate rotationally invariant representations \cite{kim2018scancontext, wang2020lidariris} and b) Descriptions based on the use of Deep Neural Networks, either operating directly on the coordinates of the points \cite{uy2018pointnetvlad} or on the projection of the points to image coordinates \cite{chen2022overlapnet}. 

\vspace{0.5cm}
In this manuscript a technique for the robust description of scenes captured by a LiDAR sensor based on the use of a Deep Neural Network is presented. Several improvements and modifications are proposed starting from the basis of several recent architectures. As a result, the proposed network is able to outperform all other existing methods in the context of place recognition. In summary, the main contributions of this paper are:
    \begin{itemize}
        \item MinkUNeXt: A new 3D Sparse Convolutional Neural Network for Place-Recognition. It is the first approach of a U-Net architecture for point cloud embedding and place-recognition. The architecture has been specifically developed and optimized for this problem. In addition, substantial improvements have been achieved both in terms of macro and micro design.
        \item The definition of a new residual block: the 3D MinkNext Block, which is entirely composed of 3D sparse convolutions and surpasses the performance of ResNet Blocks. It follows the philosophy proposed by ConvNeXt \cite{liu2022convnet}, which uses standard convolutions and was originally proposed for image classification, semantic segmentation and object detection.
    \end{itemize}

\vspace{0.5cm}

As a result, the proposed topology is able to surpass significantly the current state of the art of point cloud place-recognition in terms of average recall at 1 (AR@1) and average recall at 1\% (AR@1\%), when compared to the most relevant methods in the literature.

\vspace{0.5cm}
The rest of the paper is organized as follows: Next, Section~\ref{sota} deepens in the state of the art in relation with the use of Deep Neural Networks for the description of the structure of point clouds. After that, Section~\ref{proposed_method} defines in detail the proposed architecture to describe the point clouds. Next, Section~\ref{experiments} describes the datasets, experiments and results. Finally, Section~\ref{conclusion} presents the main conclusions.

\begin{figure*}[t]
\centering
\includegraphics[width=\textwidth]{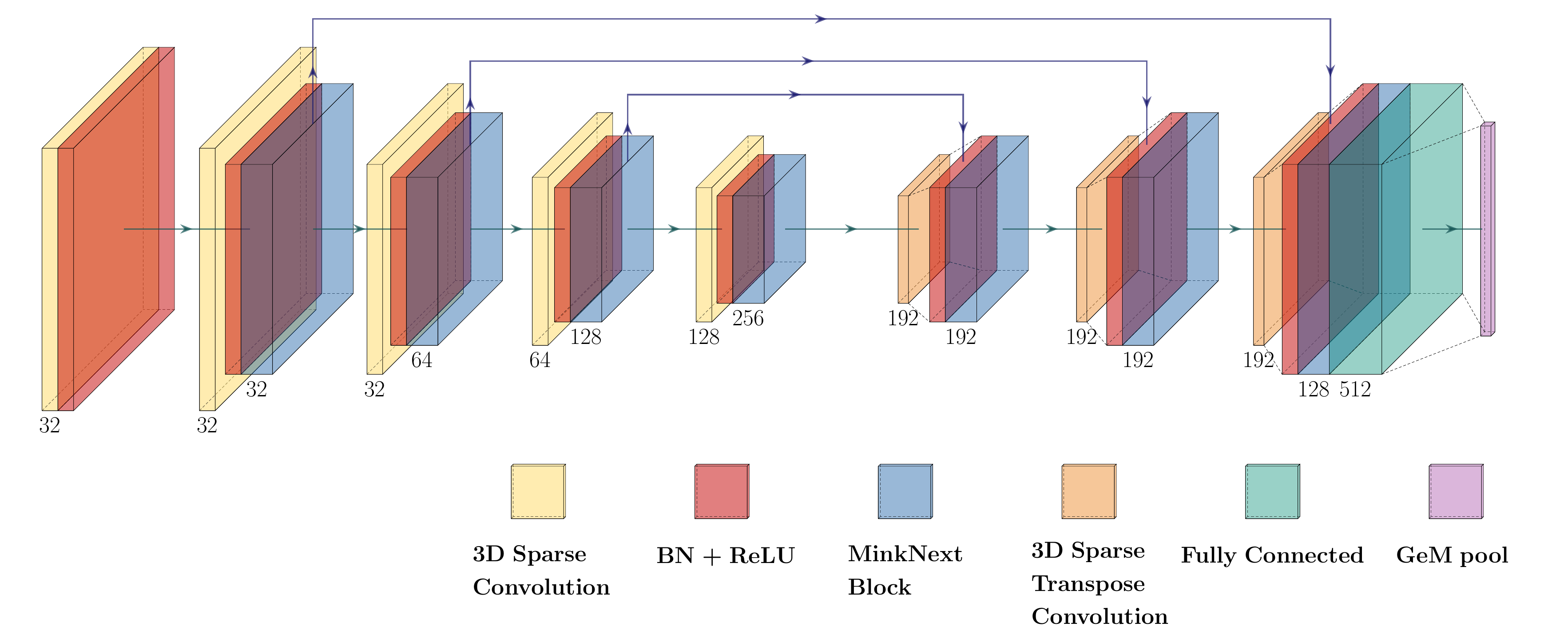}
\caption{This diagram shows the architecture of the proposed MinkUNeXt, which is based on a semantic segmentation network (U-Net) modified and enhanced to perform place-recognition from point clouds. }
\label{fig_minkunext}
\end{figure*}

\section{State of the art}\label{sota}

This section offers a comprehensive overview of the current state-of-the-art in place recognition, specifically exploring the utilization of Deep Neural Networks with point cloud data. Many applications have emerged that concentrate on place recognition based on point clouds. In this section the methods are presented chronologically. In addition, in this manuscript a comparison of the main results achieved by the most relevant architectures is provided. In this context, the first approach to this task was tackled in \cite{uy2018pointnetvlad} with PointNetVLAD, a network model based on PointNet \cite{qi2017pointnet} for feature extraction followed by a NetVLAD layer for feature aggregation. The point clouds taken as input by this type of architectures do not need to be sorted, as they use symmetric functions such as Multi Layer Perceptron (MLP) or Fully Connected layers. Next, a similar approach, named LPD-Net \cite{liu2019lpd} improved the state of the art by incorporating a local feature extraction block at the beginning of the network and a subsequent graph-based neighbourhood aggregation.

\vspace{0.5cm}
After that, the MinkLoc3D architecture emerged \cite{komorowski2021minkloc3d}. It is based on a Feature Pyramid Network (FPN) with Sparse Convolutions  for feature extraction \cite{choy20194d}, followed by a Generalized Mean Pooling (GeM) for the aggregation of the features into a single vector \cite{radenovic2018fine}. At that time, the MinkLoc3D architecture marked a significant milestone, as it significantly surpassed the existing state-of-the-art methods and also demonstrated that the use of 3D convolutional layers was a good choice for feature extraction from point clouds. Unlike previous network typologies, when using 3D convolutions, they do require a sorted point cloud as input, where the spatial relationships between points are preserved. The same situation occurs in an analogous way with images, where 2D convolutions have proven to be very efficient in feature extraction thanks to the neighbourhood relationships between pixels. In this sense, some 2D architectures have also emerged taking as input the projected point cloud into a spherical image (OverlapNet \cite{chen2022overlapnet}). Other works, such as \cite{yin2018locnet} propose creating a rotation-invariant handcrafted image: from a polar coordinate representation of the point cloud, the 2D distance between consecutive points belonging to the same elevation angle (ring) is computed and then, a histogram per ring is obtained generating a 2D handcrafted codification of the point cloud. 
 
\vspace{0.5cm}

In addition, both monocular images and point clouds are used simultaneously by some architectures (MinkLoc++ \cite{komorowski2021minkloc++}, PIC-Net \cite{lu2020pic}). In this case, both architectures are formed by two branches, processing independently the image and the point cloud. Each branch results in a feature vector and both vectors are finally aggregated into a single vector by a pooling process. Alternatively, each point can be associated with a feature corresponding to the RGB value of the image \cite{song2016}. This requires a precise calibration of the camera-LiDAR system. Otherwise, some authors propose to use the relative intensity returned by each LiDAR ray, referred to as MinkLoc-SI \cite{zywanowski2021minkloc3dSI}.

\vspace{0.5cm}
The DAGC architecture \cite{sun2020dagc} was the first to introduce self-attention layers \cite{vaswani2017attention} for point cloud feature extraction to perform place recognition. Later, other authors continued the use of attention layers, obtaining results close to the state of the art. In this sense, NDT-Transformer was presented \cite{zhou2021ndt}, a network model based on 3 Transformer Encoders that takes as input a modified point cloud by using a Normal Distribution Transform (NDT). This approach preserves the geometrical shape of the point cloud while decreasing the memory complexity. 

\vspace{0.5cm}
Simultaneously, PPT-Net \cite{hui2021ppt}, a Transformer with a pyramidal distribution followed by a NetVLAD layer, emerged. Based on a similar idea, SOE-Net \cite{xia2021soe} extracts local features using a series of MLPs and subsequently, it applies attention layers in the aggregation of those features. In addition, the Retriever \cite{wiesmann2022retriever} network also introduces self-attention layers within an autoencoder to perform local feature aggregation. Besides, looking for efficiency and the use of these architectures in real localization systems (which must work in real time), SVT-Net, an efficient Sparse Voxel Transformer based on sparse convolutional layers for feature extraction, was presented in \cite{fan2022svt}. 

\vspace{0.5cm}
Furthermore, HiTPR \cite{hou2022hitpr} employs Farthest Point Sampling \cite{qi2017pointnet++} to reduce the dimensionality of the input cloud while preserving its original topological information. In addition, this work introduces a Transformer block for short-range local feature extraction and an additional Transformer block for extracting global information over long distances. The mentioned Transformer-based approaches presented similar results to those found in the state of the art. However, the presentation of TransLoc3D \cite{xu2021transloc3d} constituted a significant advance. It is a network model also based on sparse convolutions but unlike other proposals, it extracts features at different scales in parallel by means of convolutional layers with different kernel size. In addition, it also employs ECA (Efficient Channel Attention) layers \cite{wang2020eca} in order to interact local features from different channels. This type of layers are also used by MinkLoc3Dv2 \cite{komorowski2022improving}, an architecture based on MinkLoc3D \cite{komorowski2021minkloc3d}. MinkLoc3Dv2 includes the use of ECAs with an increased number of planes or channels (depth of the convolution matrices). To date, this network architecture shows the best results in terms of average recall at 1 (AR@1) in the Oxford RobotCar Dataset \cite{maddern2017OxfordRobotCar}, partly due to the loss function they introduce in the training process and the high batch size with which they train. 

\vspace{0.5cm}
Finally, the best result in terms of average recall at 1\% (AR@1\%) was obtained by KPPR \cite{wiesmann2022kppr}, a network model based on Flexible and Deformable Convolutions (KPConv \cite{thomas2019kpconv}). However, Minkloc3Dv2 is still ahead in terms of average recall at 1 (AR@1), which is a more demanding metric. Additional architectures have been proposed to date, making other types of contributions such as rotation invariance E$^{2}$PN-GeM \cite{lin2023se} and RPR-Net \cite{fan2022rpr} or inference efficiency EPC-Net \cite{hui2022efficient} and BPT \cite{hou2023bpt}.

\vspace{0.5cm}

This paper presents MinkUNeXt, an architecture based on MinkUNet \cite{choy20194d} modified and enhanced to perform place-recognition from point clouds. It is an encoder-decoder architecture entirely based on the proposed 3D MinkNeXt Block, a residual block composed of 3D sparse convolutions that follows the philosophy proposed by ConvNeXt \cite{liu2022convnet}. The feature extraction is performed by the U-Net encoder-decoder and the feature aggregation of those features into a single descriptor is carried out by a Generalized Mean Pooling (GeM) \cite{radenovic2018GeM}. The proposed architecture demonstrates that it is possible to surpass the current state of the art by only relying on conventional 3D sparse convolutions without making use of more complex and sophisticated frameworks such as Transformers, Attention-Layers or Deformable Convolutions. In this way, this paper shows that the proposed architecture outputs results which are superior to those found in the literature while maintaining the efficiency, scalability and performance.

\section{MinkUNeXt: global point cloud descriptor
for place recognition}\label{proposed_method}

Place recognition from point clouds can be approached as an embedding task. For this purpose, it is desirable to have an architecture capable of extracting the more descriptive features of the scene and, in addition, aggregating them into a single vector that most generally describes the information present in the scene. The present work presents a pioneering solution that employs a U-Net architecture \cite{ronneberger2015unet} in the context of place recognition. Most architectures resembling U-Net were originally designed for semantic segmentation, where the goal is to assign a category to each pixel of an input image, or in this case, to each point of the input point cloud. However, the encoder-decoder topology of a U-Net is also capable to extract and fuse relevant features from the scene as will be shown in the experimental section. 

\subsection{Global Architecture}

The proposed model is fed by a point cloud given as an unordered set of 3D coordinates $P = \{(x_i, y_i, z_i)\}$. This point cloud is quantized into a sparse tensor, which is a high-dimensional extension of a sparse matrix where non-zero elements are represented as a set of indices $C$ (coordinates) and associated values (or features) $F$. Some papers \cite{liu2019lpd, zhou2021ndt} propose to employ as feature some handcrafted attributes such as the vertical component of the normal vector, height variance, change of curvature or just the value of the coordinates. Others \cite{komorowski2021minkloc3d, komorowski2022improving} prefer initializating each coordinate's feature to one, i.e., the first convolution (stem) will only take as input features `ones' for the non-empty voxels. This idea is also taken in the present paper, where the input data $\hat{P} = \{(\hat{x}_i, \hat{y}_i, \hat{z}_i, 1)\}$ is conformed by $C$, a set of 3D quantized coordinates and $F$, a vector of `ones' whose length is equal to the number of quantized points.

\vspace{0.5cm}

The global architecture is represented in Fig. \ref{fig_minkunext}.
The encoder of the network consists of five 3D Sparse Convolutions (coloured in yellow). Among them, the stem is the first convolution and it preserves the input dimension of the point cloud since its stride is fixed to 1 and the kernel size is 5. While each of the following four convolutions gradually decrease the spatial dimension, the receptive field increases since successive convolutional layers capture larger and larger patterns by combining information from previous layers. Each of those convolutions downsample its input dimension by 2 since they employ both kernel size and stride of 2. After the encoder, the dimension of the input point cloud is downsampled by 32.

\vspace{0.5cm}

In a common U-Net the decoder is composed of four 3D Sparse Transpose Convolutions that upsample the spatial dimension by 2, progressively reconstructing the input cloud. However, in this architecture it is proposed to partially reconstruct the input point cloud by only applying three transpose convolutions (coloured in orange), since our purpose is point cloud embedding and not semantic segmentation. Subsection \ref{ablation_study} will justify that features extracted with only three transpose convolutions are more robust for understanding the overall context of the scene. Furthermore, a Batch Normalization and a ReLU activation function (coloured in red) are applied after all the convolutions, which helps in stabilizing the training process. In addition, in this architecture it is proposed to employ the presented Residual MinkNeXt Block (coloured in blue) instead of the common ResNet Block after each ReLU (without taking into account the one corresponding to the stem). This kind of residual blocks provide a direct path for gradients to flow through the network, reducing overfitting and boosting the generalization capabilities on unseen data. In this architecture, it is also used to increase the number of features maps as it will further detailed in the following subsection \ref{subsec:minknext_block}.

\vspace{0.5cm}

The U-Net architecture is characterized for having skip connections between the encoder and the decoder. On the one hand, the encoder would capture features at different spatial scales, from fine details (low-level) to more global structures (high-level) present in point clouds. On the other hand, thanks to the skip connections, the decoder would fuse the low-level and high-level features. After that, a Fully Connected Layer is added since it outputs features have been proven to perform robustly against viewpoint changes in visual place recognition \cite{sunderhauf2016}. Furthermore, this Fully Connected Layer is also employed to extend the feature maps up to a dimensionality of 512. Subsequently, the points descriptors that conform that feature map are aggregated into a single global descriptor by a Generalized Mean Pooling (GeM) \cite{radenovic2018GeM}.

\subsection{Residual Block Architecture} \label{subsec:minknext_block}

\begin{figure}[h]
\centering
\includegraphics[width=0.4\textwidth]{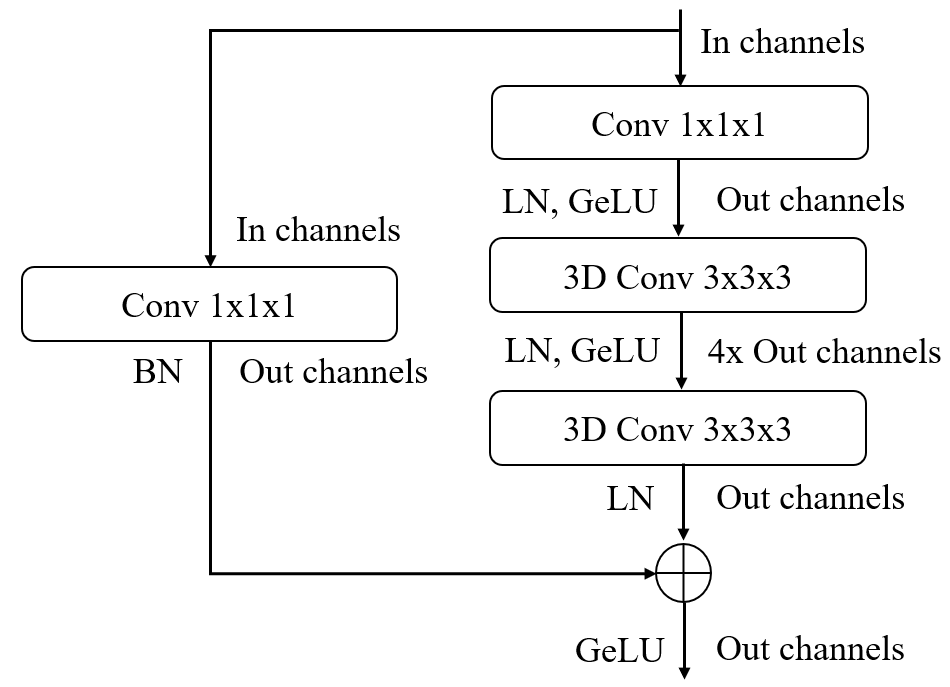}
\caption{This diagram shows the proposed MinkNeXt Block. This residual block is an essential part of the global network, since increases the number of feature maps through an inverted bottleneck.}
\label{fig:minknext_block}
\end{figure}

As mentioned before, in this paper both a global and a residual block architecture are proposed. In this sense, a new residual block is designed (Fig. \ref{fig:minknext_block}) which is entirely composed of 3D Sparse Convolutions and follows the philosophy proposed by
ConvNeXt \cite{liu2022convnet}, surpassing the performance of ResNet Blocks. We have named this block MinkNeXt, since it takes advantage of the ResNet blocks and is fully implemented in Minkowski Engine \cite{choy20194d}.

\vspace{0.5cm}

In the global architecture (Fig. \ref{fig_minkunext}), the proposed residual block appears in blue colour after each ReLU activation function (except for the one corresponding to the stem). Since the residual block is generally employed to increase the number of features maps, the stem of the residual block is formed by a 1x1x1 convolution that widens the input dimension to the output channels size.  After that, an inverted bottleneck is applied by expanding the dimension four times and then reducing it again to the output dimension through two 3D Sparse Convolutions. This inverted bottleneck was originally proposed by MobileNetV2 \cite{sandler2018mobilenetv2} and nowadays, it is an important design in every Transformer block. In addition, a 1x1x1 Convolution in the residual connection is also applied when the input and the output dimensions differ. 

\vspace{0.5cm}

The activation function employed in this block is the Gaussian Error Linear Unit
(GeLU) \cite{hendrycks2016gelu}, which is smoother than ReLU and is utilized in the most advanced Transformers. Finally, the normalization is carried out by LayerNorms \cite{ba2016layernorm} in the main stream of the block and by BatchNorms \cite{ioffe2017batchnorm} in the residual connection.

\section{Experiments} \label{experiments}

This section describes the datasets (Subsection \ref{datasets}), the labelling (Subsection \ref{subsec:labelling}) and the training and evaluation of the proposed architecture (Subsection \ref{training}). Later, the implementation details are described in Subsection \ref{implementation_details}. Subsequently, in Subsection \ref{ablation_study}, we present an ablation study of the designing steps carried out to obtain the final architecture. Finally, the main results are compared with other approaches in the literature in Subsection \ref{sota_results}.


\subsection{Datasets}
\label{datasets}


In order to train and evaluate the proposed method, the datasets and evaluation protocols introduced in \cite{uy2018pointnetvlad} have been used. This is a common framework employed and respected by a large number of studies that is used to compare different proposals that address the place recognition task using point clouds. The benchmark consists of 2 datasets and 4 different environments:

\begin{itemize}
    \item \textbf{The Oxford RobotCar Dataset \cite{maddern2017OxfordRobotCar}}. This dataset is generated using some SICK LMS-151 2D sensors mounted on a car. The dataset covers a 10 km trajectory along the city of Oxford. In total, 44 sequences of the same trajectory which are geographically divided into training (70\%) and test (30\%) are used. This results in 21,711 training submaps and 3,030 test submaps. 
    
    
    \item \textbf{The In-house Dataset \cite{uy2018pointnetvlad}}. This dataset consists of three different environments: a University Sector (U.S.), a Residential Area (R.A.), and a Business District (B.D.). These datasets are captured using a Velodyne-64 LiDAR mounted on a motorized vehicle that covers each of the three regions. The paths lengths are 10 km, 8 km and 5 km respectively. It is conformed by 5 different sequences from each of the U.S., R.A. and B.D. regions, which were captured at different times. In addition, each U.S. and R.A. sequence are geographically divided into train and test. While the B.D. environment is only used for testing. 

\end{itemize}

\vspace{0.5cm}
In both datasets, the LiDAR scans are taken at regular intervals of 12.5 m and 25 m for the training and test set, respectively. Also, both datasets are formed by a number of submaps. Each submap is constructed by capturing LiDAR scans consecutively along 20 m. Next, the scans are registered in a common frame and further processed to create a consistent submap. Each of these training and test submaps are filtered by removing the ground plane and also regularly sampled by a voxel grid filter in order to reduce its size to 4096 points. The XYZ coordinates of the points that constitute each point cloud are then shifted and scaled in order to obtain a point distribution with zero mean in the [-1, 1] range for each coordinate.

\subsection{Labelling and similarity} \label{subsec:labelling}

Each submap in the dataset is tagged with the UTM coordinates of its respective centroid. This constitutes the identifier of each submap and is later used during the training and evaluation of the network. 
Next, we define the similarity between the submaps in the datasets. This concept is generally denoted as labelling in the literature and it is important because it is necessary to feed the model with structurally similar submaps captured from the same place and structurally dissimilar submaps from different places. In this sense, most of the proposed labelling protocols are based on the Euclidean distance of the UTM coordinates from which point clouds are captured (two point clouds are considered structurally similar if they are captured within a distance $p$ and structurally different if they are taken from a distance greater than $n$ where $p < n$). This procedure, of course, is a coarse approximation that assumes that submaps captured from the same area will possess a similar structure. However, it is a simple but effective manner of labelling the training data. In this paper, this method is adopted with $p = 10 m$ and $n = 50 m$ as in the majority of the referred manuscripts. Authors, have also proposed other methods for similarity labelling in the context of place-recognition. For example, \cite{chen2022overlapnet} proposes to use the overlap between point clouds as an alternative method for labelling similar and dissimilar point clouds. In order to compute the overlap between two point clouds (i.e. submaps) a precise registration must be carried out, which hinders the application of this technique to large datasets. 

\subsection{Training and evaluation}\label{training}
As for the training and evaluation of the proposed method, the two evaluation protocols established in \cite{uy2018pointnetvlad} have been followed: 
\begin{itemize}
    \item The first, baseline protocol, consists in training the model only with the Oxford training data and evaluating with the Oxford and In-house (U.S., R.A. and B.D.) test data.
    \item The second, refined protocol, consists in training with Oxford and In-house (U.S., R.A.) training data and evaluating with the Oxford and In-house (U.S., R.A. and B.D.) test data.
\end{itemize}


Table \ref{tab:protocols} summarizes the number of training and testing submaps corresponding to each dataset and each of the protocols defined above. The assessment of the LiDAR-based place recognition descriptors is carried out by means of the recall rate at top-K candidates. Following the most common evaluation methods (as in the manuscripts cited in Section \ref{sota}), the average recall at 1 (AR@1) and average recall at 1\% (AR@1\%) are used in order to ease the comparison with other techniques. We start with a ``query submap'' formed by a point cloud which is taken from the test dataset and point clouds submaps from different traversals that cover the same region from the map. Each query submap is processed by the network and it outputs, as a result, a descriptor vector that codifies its appearance. This descriptor is referred to as the ``query descriptor''. Next, the query descriptor is compared to all the descriptors in the map. The point cloud in the database that minimizes the distance is selected. Finally, the place recognition is considered to be successful if the query and the retrieved point cloud are within an Euclidean distance of 25 m.

\begin{table}[t]
\caption{Number of training and testing submaps for the baseline and refined protocols.}
\label{tab:protocols}
\centering
\begin{tabular}{ccccc}
\hline
         & \multicolumn{2}{c}{\textbf{Baseline Protocol}} & \multicolumn{2}{c}{\textbf{Refined Protocol}} \\
         & Training            & Test            & Training           & Test           \\ \hline
Oxford   & 21.7k               & 3.0k            & 21.7k              & 3.0k           \\
In-house & -                   & 4.5k            & 6.7k               & 1.7k           \\ \hline
\end{tabular}
\end{table}



\subsection{Implementation details} \label{implementation_details}

In the present work the proposed model is trained following the procedure established in \cite{komorowski2022improving}. In this regard, the Truncated Smooth-AP (TSAP) loss function is employed, which tries to maximize the ranking of the positive top-k candidates:
\begin{equation}
\mathcal{L}_{TSAP} = \frac{1}{b} \sum_{q=1}^{b}(1 - AP_q) 
\label{eq:TSAP}
\end{equation}

Where $b$ is the batch size and $AP_{q}$ is the smooth average precision:
\begin{equation}
AP_q = \frac{1}{\lvert P \rvert}\sum_{i \in P} \frac{1 + \sum_{j \in P, j \neq i} G(d(q,i) - d(q,j); \tau)}{1 + \sum_{j \in \Omega, j \neq i} G(d(q,i) - d(q,j); \tau)}
\label{eq:AP_q}
\end{equation}

Given a query point cloud $q$, the average precision $AP_q$ is computed from the set of $k$ closest candidates $P$ (positives) and the set of  all positives and negatives $\Omega$. Also, the function $G$ constitutes a Sigmoid function $G(x;\tau) = \left(1 + \exp\left(-\frac{x}{\tau}\right)\right)^{-1}$ with a parameter $\tau$ that controls the sharpness. The term $d(q,i)$ represents the Euclidean distance between the descriptor of a query point cloud $q$ and the $i$-th point cloud. The numerator represents a soft ranking of a positive point $i$ among the top $k$ positives (where $k=4$), while the denominator represents a soft ranking of a positive point $i$ among all other positives and negatives.

\vspace{0.5cm}


For the correct performance of this type of loss function, it is necessary to train with a high batch size, specifically a size of 2048 has been used with 400 and 500 training epochs for the baseline and refined protocol, respectively. The optimizer used to minimize the loss function is Adam with an Initial Learning Rate of 1e-3 and it is divided by 10 in the epochs given by the LR scheduler steps, which are epochs 250 and 350 for baseline protocol and epochs 350 and 450 for refined protocol. Table \ref{tab:training_parameters} summarizes all the parameter values described above.

\vspace{0.5cm}

Additionally, when working with sparse convolutions, the input point clouds need to be quantized by a factor of $qs$, which is set to 0.01. Since these clouds are already normalized to [-1, 1], spatial resolutions of 200 voxels are obtained on each coordinate axis. To increase the number of training instances and reduce model overfitting, a data augmentation has been carried out by applying a random jitter of a value between [0, 0.001] individually to each point of the point cloud, a random transformation to the global point cloud with a value between [0, 0.01] and a random removal of 10\% of the points.

\vspace{0.5cm}

All experiments are carried out on a NVIDIA GeForce RTX 3090 GPU with 24 GB. Our code is publicly available on the project website \href{https://github.com/juanjo-cabrera/MinkUNeXt.git}{https://github.com/juanjo-cabrera/MinkUNeXt.git}.

\begin{table}[t]
  \centering
  \caption{Training Parameters in Baseline and Refined Protocols}
  \label{tab:training_parameters}
  \begin{tabular}{@{}lcc@{}}
    \toprule
    \textbf{Parameter}            & \textbf{Baseline} & \textbf{Refined} \\ 
    \midrule
    Batch Size ($b$)                    & 2048              & 2048             \\
    Number of Epochs              & 400               & 500              \\
    Initial Learning Rate         & $1 \times 10^{-3}$ & $1 \times 10^{-3}$ \\
    LR Scheduler Steps            & 250, 350          & 350, 450         \\
    L2 Weight Decay               & $1 \times 10^{-4}$ & $1 \times 10^{-4}$ \\
    Sigmoid Temperature ($\tau$)  & 0.01              & 0.01             \\
    Positives per Query ($k$)     & 4                 & 4                \\
    Quantization Scale ($qs$)     & 0.01              & 0.01                \\
    \bottomrule
  \end{tabular}
\end{table}

\subsection{Ablation study: From MinkUNet to MinkUNeXt} \label{ablation_study}
\begin{figure*}[t]
\centering
\includegraphics[width=\textwidth]{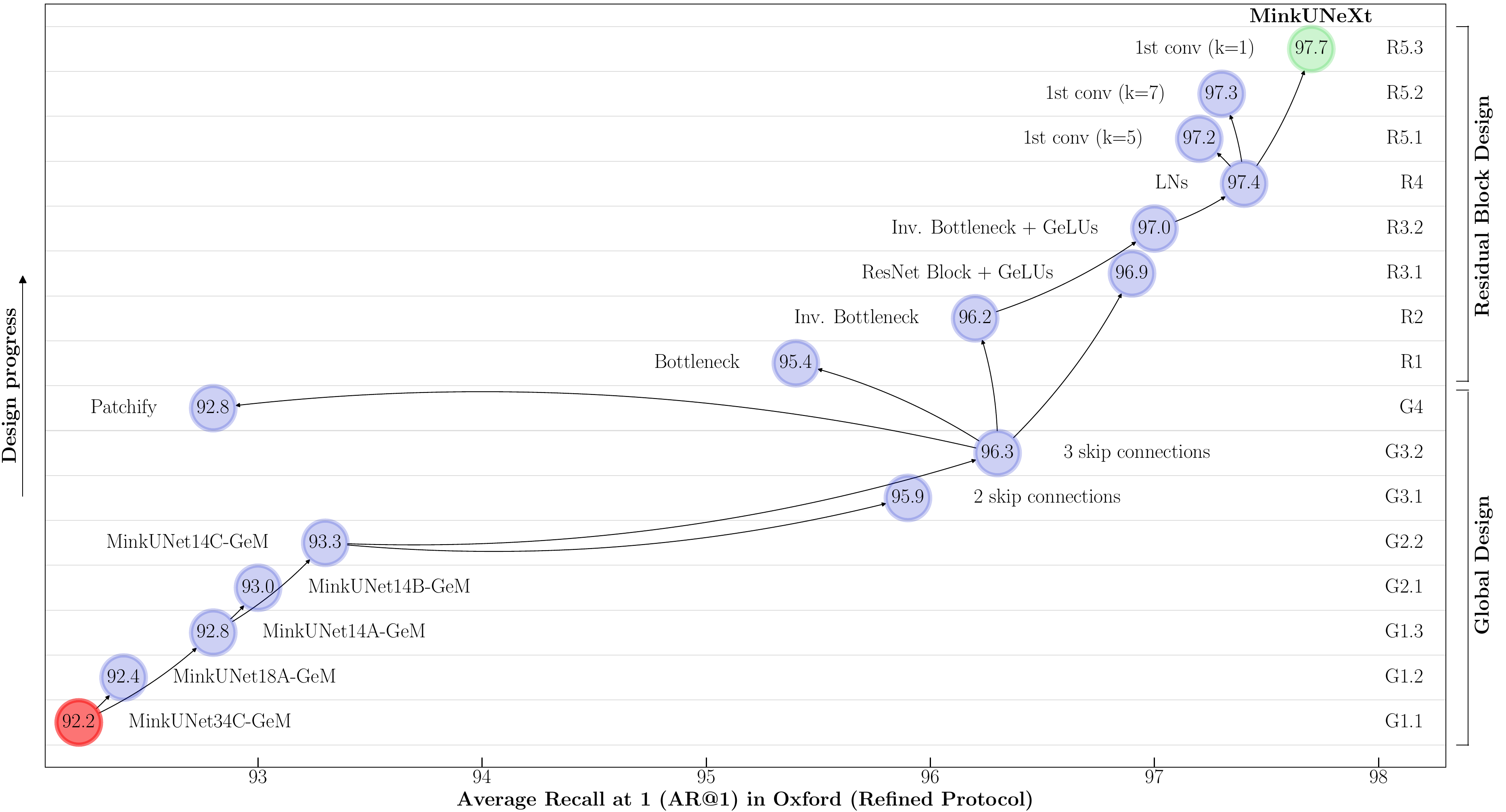}
\caption{This diagram illustrates the design progress of the proposed architecture from MinkUNet up to MinkUNeXt. All the proposed modifications are summarized in Table \ref{tab:comentarios_nodo}.}
\label{diagrama}
\end{figure*}

The design departs from the MinkUNet34C architecture \cite{choy20194d} as a baseline. Next, the series of design decisions are described. Each design step is summarized in two main subsections: (1) global design and (2) residual block design, which are included next. For every step, both the procedure and the results are presented, starting from the MinkUNet34C until obtaining the MinkUNeXt architecture. The evolution of the network and results is presented in Fig. \ref{diagrama}. Table \ref{tab:comentarios_nodo} summarizes and describes the main design steps.

\vspace{0.5 cm}

\subsubsection{Global Design}



As mentioned above, the starting point is the MinkUnet34C \cite{choy20194d} architecture and it is first modified by adding a GeM pool layer. This step is marked as G1.1 in Fig. \ref{diagrama}. The rest of the roadmap followed towards the final design is described next. Each of the design steps is classified in one of the following points: evaluating the cardinality, evaluating the number of channels, changing the number of skip connections and changing the stem to ``Patchify".

\vspace{0.5cm}
\begin{enumerate}
  \item[\textbf{G1.}] \textbf{Evaluating the cardinality.} The cardinality is defined as the number of parallel blocks, that enables the network to learn various input representations. In this sense, different cardinality configurations per residual block are assessed: (2, 3, 4, 6, 2, 2, 2, 2), (2, 2, 2, 2, 2, 2, 2, 2) and (1, 1, 1, 1, 1, 1, 1, 1), corresponding to MinkUNet34, MinkUNet18 and  MinkUNet14, respectively. These cardinality values represent the number of instances of each Residual Block that appear in blue colour in Fig. \ref{fig_minkunext}, but at this point still with ResNet Blocks. In addition, these cardinality configurations are summarized respectively in steps G1.1, G1.2 and G1.3 in Fig. \ref{diagrama}. As ilustrated in the diagram, reducing the cardinality to the minimum, with no parallel blocks, shows a better performance and allows an improvement from 92.2\% to 92.8\% in terms of average recall at 1 (AR@1). From now on, it will be used 1 as cardinality of each residual block.

  \vspace{0.5cm}
  \item[\textbf{G2.}] \textbf{Evaluating the number of channels.} The number of channels or filters correspond to the number of feature maps that the convolutional layer can learn. The number of filters corresponding to the convolutional layers of the encoder are fixed to (32, 64, 128, 256), but the number of channels of the decoder takes the following values (128, 128, 96, 96), (128, 128, 128, 128) and (192, 192, 128, 128) corresponding to MinkUNet14A, MinkUNet14B and MinkUNet14C. This number of filters of the decoder is summarized respectively in steps G1.3, G2.1 and G2.2 in Fig. \ref{diagrama}. The best result is obtained with MinkUnet14C (G2.2) with an AR@1 of 93.3\%. Thus, the number of filters of the transposed convolutions that will be adopted in the subsequent architecture variations is (192, 192, 128, 128).
  
  \vspace{0.5cm}
  \item[\textbf{G3.}] \textbf{Changing the number of skip connections.} The original U-Net is characterized by the presence of 4 skip connections between the encoder and the decoder networks. In this sense, in the present paper the performance of the network is studied when reducing the number of skip connections and removing the transpose convolutions after the last connection. Apart from the 4 skip connections already implemented in the above configurations, we have evaluated 2 and 3 skip connections corresponding to G3.1 and G3.2 in Fig. \ref{diagrama}. By reducing the number of skip connections to 3 and removing the layers after the last connection, the model shows, by far, the greatest improvement on the AR@1, increasing the results from 93.3\% to 96.3\%. As a result, only 3 skip connections will be included between the encoder and decoder.

  \vspace{0.5cm}
  \item[\textbf{G4.}] \textbf{Changing the stem to ``Patchify".} The stem refers to the first layer in the network, which performs the initial processing. In this case, the first processing is carried out by a 3D Sparse Convolution with kernel size 5 and stride 1. The term ``Patchify" refers to the act of splitting the input data into a sequence of patches. Visual Transformers \cite{dosovitskiy2020ViT} introduced this concept, originally inspired by NLP Transformers \cite{vaswani2017attention}. The Swin Transformer \cite{liu2021swin} uses as stem a non-overlapping convolution with kernel size 4 and stride 4. In this sense, these parameters are adopted for the stem in G4, but the performance of the network has decreased from 96.3\% to 92.8\% so ``Patchify" is discarded. 
\end{enumerate}

\vspace{0.5cm}

\subsubsection{Residual Block Design}

This section describes each design step from ResNet Block to the proposed MinkNeXt Block. The roadmap of the design of this residual block is divided in the following points: creating a Bottleneck in the residual block, creating an Inverted Bottleneck in the residual block, replacing ReLUs with GeLUs, substituting BN with LN and evaluating different kernel sizes.

\vspace{0.5cm}
\begin{enumerate}
    \item[\textbf{R1.}] \textbf{Creating a Bottleneck in the residual block.} A Bottleneck consists in reducing the dimensionality of the hidden layer and then expanding it to its original size using 1x1 convolutions. This modification led to worse results in the performance of the proposed architecture.
  
    \vspace{0.5cm}
    \item[\textbf{R2.}] \textbf{Creating an Inverted Bottleneck in the residual block.} Every Transformer block is characterized by an inverted bottleneck, which consists in expanding the dimensionality of the feature map of the hidden layer and then reducing it to its original size by 1x1 convolutions. In this case, 3D sparse convolutions with kernel size 3 and stride 1 are employed to create the inverted bottleneck with a hidden dimension four times wider than the input dimension. Fig. \ref{diagrama} shows that this inverted bottleneck block produces better results compared to the previous ResNet block when analysed jointly with the following modification (R3).

    \vspace{0.5cm}
    \item[\textbf{R3.}] \textbf{Replacing ReLUs with GeLUs.} The Rectified Linear Unit (ReLU) \cite{nair2010relu} is the most employed activation function over time due to its simplicity and efficiency. However, recent advanced Transformers such as Google's BERT \cite{devlin2018bert} or OpenAI's GPT-4 \cite{achiam2023gpt} employ Gaussian Error Linear Units (GeLUs) \cite{hendrycks2016gelu}, which is a smoother variant of ReLUs. Following the same philosophy, ReLUs are replaced with GeLUs in both the ResNet Block and the inverted bottleneck block, steps R3.1 and R3.2 in Fig. \ref{diagrama}, respectively. In both cases, the performance of the architecture improves, but better results are obtained with the proposed inverted bottleneck block, achieving an AR@1 of 97.0\%. In consequence, an inverted bottleneck with GeLUs will be used as residual block.

    \vspace{0.5cm}
    \item[\textbf{R4.}] \textbf{Substituting BN with LN.} Batch Normalization (BN) \cite{ioffe2017batchnorm} plays a critical role in convolutional networks by enhancing convergence and mitigating overfitting. However, BN may introduce complexities that may negatively impact the model's performance. Recently, the simpler Layer Normalization \cite{ba2016layernorm} (LN) has been successfully implemented in Transformers. Thus, BN is replaced with LN in the proposed residual block, obtaining an improvement of the model performance up to 97.4\%. As a result, Layer Normalization will be employed instead of Batch Normalization in the residual block.
    
    \vspace{0.5cm}
    \item[\textbf{R5.}] \textbf{Evaluating different kernel sizes.} Vision Transformers are characterized by employing large kernel sizes with a minimum dimension of 7. However, as shown in Fig. \ref{diagrama} (R5), the usage of smaller kernel sizes is beneficial in the present place recognition task, both in the input, hidden and last layers of the residual block. In this sense, we find the best parameter configuration with a kernel size of 1 in the first convolution and kernel sizes of 3 in the hidden and last convolutions. This leads to the final model and residual block architectures, which we have named MinkUNeXt and MinkNeXt block, respectively.
\end{enumerate}

\begin{table}[h]
  \centering
  \caption{This table summarizes all modifications proposed in the architecture design progress from MinkUNet up to MinkUNeXt. }
  \label{tab:comentarios_nodo}
  \begin{tabular}{@{}ll@{}}
    \toprule
    \textbf{ID} & \textbf{Design modifications} \\
    \midrule
    G1.1 & Cardinality: (2, 3, 4, 6, 2, 2, 2, 2) \newline $\rightarrow$ (2, 2, 2, 2, 2, 2, 2, 2) \\
    G1.2 & Cardinality: (2, 2, 2, 2, 2, 2, 2, 2) \newline $\rightarrow$ (1, 1, 1, 1, 1, 1, 1, 1) \\
    G2.1 & Decoder channels: (128, 128, 96, 96) \newline $\rightarrow$ (128, 128, 128, 128) \\
    G2.2 & Decoder channels: (128, 128, 96, 96) \newline $\rightarrow$ (192, 192, 128, 128) \\
    G3.1 & 4 skip connections $\rightarrow$ 2 skip connections \\
    G3.2 & 4 skip connections $\rightarrow$ 3 skip connections \\
    G4   & Stem (k=5, s=1 $\rightarrow$ k=4, s=4) \\
    R1   & ResNet Block $\rightarrow$ Bottleneck \\
    R2   & ResNet Block $\rightarrow$ Inv. Bottleneck \\
    R3.1 & ResNet Block with ReLUs $\rightarrow$ ResNet Block with GeLUs \\
    R3.2 & Inv. Bottleneck with ReLUs $\rightarrow$ Inv. Bottleneck with GeLUs \\
    R4   & Inv. Bottleneck with BNs $\rightarrow$ Inv. Bottleneck with LNs \\
    R5.1 & Inv. Bottleneck 1st convolution (k=3 $\rightarrow$ k=5) \\
    R5.2 & Inv. Bottleneck 1st convolution (k=3 $\rightarrow$ k=7) \\
    R5.3 & Inv. Bottleneck 1st convolution (k=3 $\rightarrow$ k=1) \\
    \bottomrule
  \end{tabular}
\end{table}

\subsection{Comparison with the state of the art}\label{sota_results}
As defined in Subsection \ref{training}, the two training and evaluation protocols established in \cite{uy2018pointnetvlad} have been followed for place recognition with the Oxford RobotCar and In-house datasets. The baseline protocol consists in training the model only with the Oxford training data and evaluating with the Oxford and In-house (U.S., R.A. and B.D.) test data. In contrast, the refined protocol consists in training with Oxford and In-house (U.S., R.A.) training data and evaluating with the Oxford and In-house (U.S., R.A. and B.D.) test data. These protocols are widely used in the literature, so that the comparison is performed on the same terms and conditions. Additionally, the comparative results shown here have been obtained from the same works that are referenced. 

\vspace{0.5cm}
Tables \ref{tab:results_baseline} and \ref{tab:results_refined} present an overview of the results with different techniques proposed in the state of the art compared to the one proposed in this paper under the same training and evaluation protocols (baseline and refined), in terms of average recall at 1 (AR@1) and average recall at 1\% (AR@1\%). Each column presents the results obtained with each of the datasets, whereas the last two columns present the mean results.

\vspace{0.5cm}
\subsubsection{Results with the Baseline Protocol}
Table \ref{tab:results_baseline} presents the results of several methods in terms of average recall at 1 (AR@1) and average recall at 1\% (AR@1\%). It can be observed that, PointNetVLAD established the starting point for place-recognition from point clouds with the Oxford Robotcar and the In-house dataset. PCAN slightly outperforms PointNetVLAD on most datasets. BPT stands out with really competitive results, especially in Oxford and U.S. RPR-Net outperforms BPT in U.S, R.A and B.D., showing better generalization capabilities. Some works, such as DAGC and Retriever, do not provide AR@1 results for all datasets. However, they presented AR@1\% results which show a performance better than PCAN, but worse than BPT. Futhermore, LPD-Net, HiTPR, EPC-Net and E$^{2}$PN-GeM show similar, but good results across multiple scenarios. SOE-Net, only provides AR@1\% results which are really promising as they are close to MinkLoc3D, the first architecture to exceed 90\% in AR@1 with the Oxford dataset. Moreover, HiBi-Net, PPT-Net and SVT-Net show slightly higher performance, specifically for the In-house dataset. TransLoc3D takes a step forward with the best result so far in Oxford and solid performance in the other scenarios, and its improved version MinkLoc3Dv2 outperforms the rest of the architectures. In addition, KPPR also shows a remarkable performance, but only presented average recall at 1\% results in the case of U.S., R.A., B.D. 

\vspace{0.5cm}
Finally, the proposed architecture, MinkUNeXt, demonstrates superior performance in terms of AR@1 and AR@1\% on Oxford. It outperforms all of the existing methods with a 97.5\% in AR@1 and 99.3\% in AR@1\%. However, the performance slightly decreases when the model is tested in U.S., R.A. and B.D. It should be highlighted that the Oxford dataset and the three in-house datasets were obtained using LiDARs that exhibit different characteristics, such as number of channels or spatial resolution. The Oxford dataset is captured with various SICK LMS-151 2D and the In-house dataset with a 64 channel Velodyne. Moreover, the submaps within the Oxford dataset contain scenes that are entirely urban, characterized by densely built environments with a more compact structure. In contrast, the scenarios present in the in-house dataset are considerably more open, with fewer obstructions and a more dispersed arrangement of urban elements. This difference in the nature of the captured scenes can significantly influence the results and performance of the model on each dataset.

\subsubsection{Results with the Refined Protocol}

\vspace{0.5cm}

As for the performances of the models when training with the refined protocol (Table \ref{tab:results_refined}), PointNetVLAD introduced the starting reference point as well, surprisingly achieving a good performance in U.S. R.A. and B.D. despite the simplicity of the network architecture. PCAN and DAGC presented similar results to PointNetVLAD for the In-house dataset, but especially better in Oxford. In contrast, LPD-Net and SOE-Net show substantially better performance in all metrics and datasets. MinkLoc3D also manages to exceed 90\% on average recall at 1 (AR@1) in Oxford and generally performs well in all metrics and sets. PPT-Net does not provide values for average recall at 1 (AR@1), but shows a promising performance on average recall at 1\% (AR@1\%). Furthermore, SVT-Net stands out especially in U.S., R.A. and B.D. In addition, TransLoc3D achieves good results in all metrics, being one of the best methods overall. MinkLoc3Dv2 boasted the best results in the state of the art so far, showing improvements over MinkLoc3D.

\vspace{0.5cm}
Finally, the proposed MinkUNeXt model shows considerable improvements in average recall at 1 (AR@1) and average recall at 1\% (AR@1\%) for all the scenarios obtaining the best results of the state of the art so far. The average recall at 1 (AR@1) metric on Oxford dataset is 97.7\% and outperforms the runner-up MinkLoc3Dv2 by 0.8 p.p. On the R.A., B.D. scenarios it surpasses MinkLoc3Dv2 by 0.1 to 1.1 p.p. Nevertheless, slightly worse results (0.3 p.p.) are obtained with this metric in the U.S. dataset. Regarding the results in terms of AR@1\% for the refined protocol, there was little room for improvement. However, the results on Oxford are improved by 0.2 p.p. to reach 99.3\%, on R.A. by 0.5 p.p. to reach 99.9\% and on B.D. by 0.1 p.p. to reach 97.7\%. In addition, although the model previously output slightly worse results for U.S. in terms of AR@1, the performance of the network in the AR@1\% metric is equal to the best previous result in the state-of-the-art with a value of 99.9\%. The mean AR@1 and AR@1\% over all 4 datasets improves by 0.4\% and 0.2\%, respectively. To conclude, training the MinkUNeXt with the refined protocol overcomes the generalization difficulties presented when training with the baseline protocol, since the model adapts to both LiDAR characteristics.


\begin{table*}
\centering
\caption{EVALUATION RESULTS IN TERMS OF AVERAGE RECALL AT 1 (AR@1) AND AT 1\% (AR@1\%) OF PLACE RECOGNITION METHODS TRAINED USING THE BASELINE PROTOCOL.}
\label{tab:results_baseline}
\begin{tabular}{lcccccccc|cc}
\toprule
& \multicolumn{2}{c}{Oxford} & \multicolumn{2}{c}{U.S.} & \multicolumn{2}{c}{R.A.} & \multicolumn{2}{c}{B.D.} & \multicolumn{2}{c}{Mean} \\
\cmidrule(lr){2-3} \cmidrule(lr){4-5} \cmidrule(lr){6-7} \cmidrule(lr){8-9} \cmidrule(lr){10-11}
Method & AR@1 & AR@1\% & AR@1 & AR@1\% & AR@1 & AR@1\% & AR@1 & AR@1\% & AR@1 & AR@1\% \\
\midrule
PointNetVLAD \cite{uy2018pointnetvlad} & 62.8 & 80.3 & 63.2 & 72.6 & 56.1 & 60.3 & 57.2 & 65.3 & 59.8 & 69.6 \\
PCAN \cite{zhang2019pcan} & 69.1 & 83.8 & 62.4 & 79.1 & 56.9 & 71.2 & 58.1 & 66.8 & 61.6 & 75.2 \\
DAGC \cite{sun2020dagc} & - & 87.5 & - & 83.5 & - & 75.7 & - & 71.2 & - & 79.5 \\
BPT \cite{hou2023bpt} & 85.7 & 93.3 & 80.5 & 89.3 & 77.4 & 86.6 & 74.1 & 78.5 & 79.4 & 86.9\\
Retriever \cite{wiesmann2022retriever} & - & 91.9 & - & 91.9 & - & 87.4 & - & 85.5 & - & 89.2 \\
RPR-Net \cite{fan2022rpr} & 81.0 & 92.2 & 83.2 & 94.5 & 83.3 & 91.3 & 80.4 & 86.4 & 82.0 & 91.1 \\
LPD-Net \cite{liu2019lpd} & 86.3 & 94.9 & 87.0 & 96.0 & 83.1 & 90.5 & 82.5 & 89.1 & 84.7 & 92.6 \\
HiTPR \cite{hou2022hitpr} & 87.8 & 94.6 & 86.0 & 94.0 & 81.3 & 89.1 & 81.8 & 88.3 & 84.2 & 91.5\\
EPC-Net \cite{hui2022efficient} & 86.2 & 94.7 & - & 96.5 & - & 88.6 & - & 84.9 & - & 91.2 \\
E$^{2}$PN-GeM \cite{lin2023se} & 84.8 & 93.2 & 88.1 & 95.3 & 83.7 & 90.5 & 83.3 & 87.7 & 85.0 & 91.7 \\
SOE-Net \cite{xia2021soe} & - & 96.4 & - & 93.2 & - & 91.5 & - & 88.5 & - & 92.4 \\
MinkLoc3D \cite{komorowski2021minkloc3d} & 93.0 & 97.9 & 86.7 & 95.0 & 80.4 & 91.2 & 81.5 & 88.5 & 85.4 & 93.2 \\
HiBi-Net \cite{shu2023hierarchical} & 87.5 & 95.1 & 87.8 & - & 85.8 & - & 83.0 & - & 86.0 & - \\
NDT-Transformer \cite{zhou2021ndt} & 93.8 & 97.7 & - & - & - & - & - & - & - & - \\
PPT-Net \cite{hui2021ppt} & 93.5 & 98.1 & 90.1 & 97.5 & 84.1 & 93.3 & 84.6 & 90.0 & 88.1 & 94.7 \\
SVT-Net \cite{fan2022svt} & 93.7 & 97.8 & 90.1 & 96.5 & 84.3 & 92.7 & 85.5 & 90.7 & 88.4 & 94.4 \\
TransLoc3D \cite{xu2021transloc3d} & 95.0 & 98.5 & - & 94.9 & - & 91.5 & - & 88.4 & - & 93.3 \\
MinkLoc3Dv2 \cite{komorowski2022improving} & 96.3 & 98.9 & \textbf{90.9} & 96.7 & \textbf{86.5} & 93.8 & \textbf{86.3} & 91.2 & \textbf{90.0} & 95.1 \\
KPPR \cite{wiesmann2022kppr} & 91.5 & 97.1 & - & \textbf{98.0} & - & \textbf{95.1} & - & \textbf{92.1} & - & \textbf{95.6} \\
 MinkUNeXt (ours) & \textbf{97.5} & \textbf{99.3} & 88.9 & 96.5 & 85.0 & 91.3 & 85.2 &  90.1 & 89.1 &  94.3 \\
\bottomrule
\end{tabular}
\end{table*}

\begin{table*}
\centering
\caption{EVALUATION RESULTS IN TERMS OF AVERAGE RECALL AT 1 (AR@1) AND AT 1\% (AR@1\%) OF PLACE RECOGNITION METHODS TRAINED USING THE REFINED PROTOCOL.}
\label{tab:results_refined}
\begin{tabular}{lcccccccc|cc}
\toprule
& \multicolumn{2}{c}{Oxford} & \multicolumn{2}{c}{U.S.} & \multicolumn{2}{c}{R.A.} & \multicolumn{2}{c}{B.D.} & \multicolumn{2}{c}{Mean} \\
\cmidrule(lr){2-3} \cmidrule(lr){4-5} \cmidrule(lr){6-7} \cmidrule(lr){8-9} \cmidrule(lr){10-11}
Method & AR@1 & AR@1\% & AR@1 & AR@1\% & AR@1 & AR@1\% & AR@1 & AR@1\% & AR@1 & AR@1\% \\
\midrule
PointNetVLAD \cite{uy2018pointnetvlad} & 63.3 & 80.1& 86.1  & 94.5 & 82.7 & 93.1  & 80.1 & 86.5 & 78.0 & 88.6 \\
PCAN \cite{zhang2019pcan} & 70.7 & 86.4 & 83.7 & 94.1 & 82.5 & 92.5 & 80.3 & 87.0 & 79.3 & 90.0 \\
DAGC \cite{sun2020dagc} & 71.5 & 87.8 & 86.3 & 94.3 & 82.8 & 93.4 & 81.3 & 88.5 & 80.5 & 91.0 \\
LPD-Net \cite{liu2019lpd} & 86.6 & 94.9 & 94.4 & 98.9 & 90.8 & 96.4 & 90.8 & 94.4 & 90.7 & 96.2 \\
SOE-Net \cite{xia2021soe} & 89.3 & 96.4 & 91.8 & 97.7 & 90.2 & 95.9 & 89.0 & 92.6 & 90.1 & 95.7 \\
MinkLoc3D \cite{komorowski2021minkloc3d} & 94.8 & 98.5 & 97.2 & 99.7 & 96.7 & 99.3 & 94.0 & 96.7 & 95.7 & 98.6 \\
PPT-Net \cite{hui2021ppt} & - & 98.4 & - & 99.7 & - & 99.5 & - & 95.3 & - & 98.2 \\
SVT-Net \cite{fan2022svt} & 94.7 & 98.4 & 97.0 & \textbf{99.9} & 95.2 & 99.5 & 94.4 & 97.2 & 95.3 & 98.8 \\
TransLoc3D \cite{xu2021transloc3d} & 95.0 & 98.5 & 97.5 & 99.8 & 97.3 & 99.7 & 94.8 & 97.4 & 96.2 & 98.9 \\
MinkLoc3Dv2 \cite{komorowski2022improving} & 96.9 & 99.1 & \textbf{99.0} & 99.7 & 98.3 & 99.4 & 97.6 & \textbf{99.1} & 97.9 & 99.3\\
MinkUNeXt (ours) & \textbf{97.7} & \textbf{99.3}  & 98.7  & \textbf{99.9} & \textbf{99.4} & \textbf{99.9}   & \textbf{97.7} & 99.0  & \textbf{98.3} & \textbf{99.5}\\
\bottomrule
\end{tabular}
\end{table*}

\section{Conclusion} \label{conclusion}

This paper presents MinkUNeXt, an architecture based on MinkUNet \cite{choy20194d} exhaustively modified and enhanced to perform place-recognition based on point clouds. It is an encoder-decoder architecture entirely based on the proposed 3D MinkNeXt Block: a residual block composed of 3D sparse convolutions that follows the philosophy proposed by ConvNeXt \cite{liu2022convnet}. The feature extraction step is performed by a U-Net encoder-decoder. The feature aggregation of those features into a single descriptor is carried out by a Generalized Mean Pooling (GeM) \cite{radenovic2018GeM}. The designed architecture demonstrates that it is possible to surpass the current state of the art by only relying on conventional 3D sparse convolutions without making use of more complex and sophisticated proposals such as Transformers, Attention-Layers or Deformable Convolutions.

\vspace{0.5 cm}

The proposed network shows that the usage of a U-Net architecture for point cloud-based place recognition is beneficial, since it is able to capture both detailed and contextual information of the three-dimensional environment. The fusion of features from multiple spatial scales improves the robustness of the place recognition model, allowing it to adapt to variations in point cloud geometry and density, as well as to different scenarios.

\vspace{0.5 cm}

It should be also noted that the proposed method outputs results outperforming an already saturated state-of-the-art. In particular, the network achieved an AR@1 of 97.5\% and an AR@1\% of 99.3\% when trained with the refined protocol. Thus, there is little room for improvement and larger and more diverse scenarios are needed in order to stimulate further progress.

\vspace{0.5 cm}

Future work will consider the inclusion of visual information into the place recognition system. In this sense, we consider that it would result in a richer representation of the environment compared to the use of LiDAR with pure distance data. However, visual information is hindered by changing lighting conditions, weather and seasonal changes, which pose a great challenge. 


\section*{Acknowledgments}
The Ministry of Science, Innovation and Universities (Spain) has supported this work through ``Ayudas para la Formación de Profesorado Universitario'' (FPU21/04969). This work is also part of the project TED2021-130901B-I00, funded by MCIN/AEI/10.13039501100011033 and the European Union ``NextGenerationEU”/PRTR, and of the project PROMETEO/2021/075 funded by Generalitat Valenciana (Spain). In addition, this research was sponsored by national funds (Portugal) through FCT - Fundação para a Ciência e a Tecnologia, under project LA/P/0079/2020, DOI: 10.54499/LA/P/0079/2020. This work was further funded in the scope of the projects E-Forest—Multi-agent Autonomous Electric Robotic Forest Management Framework, ref. POCI-01-0247-FEDER-047104 and F4F—Forest for Future, ref. CENTRO-08-5864-FSE-000031, co-financed by European Funds through the programs Compete 2020 and Portugal 2020.

\vspace{1cm}



 
%

\bibliographystyle{IEEEtran}
\bibliography{bibliography}


 





\end{document}